\RequirePackage{amsmath}

\documentclass[runningheads]{llncs}
\usepackage{makeidx}
\usepackage{graphicx}
\usepackage{amsmath,amssymb} 
\usepackage{color}

\usepackage{etoolbox}
\usepackage{float}
\usepackage{subcaption}
\usepackage[hidelinks]{hyperref}
\usepackage{multirow}
\usepackage{times}
\newbool{draft}
\booltrue{draft}
\boolfalse{draft}
\ifbool{draft}{\usepackage{geometry}}{}
\ifbool{draft}{\geometry{a4paper,total={150mm, 260mm},left=10mm,top=20mm}}{}
\ifbool{draft}{\usepackage{setspace}}{}
\ifbool{draft}{\doublespacing}{}
\ifbool{draft}{\usepackage{fancyhdr}}{}
\ifbool{draft}{\usepackage{datetime}}{}

\newcommand{\etal}{\textit{et al}. } 
\DeclareMathOperator{\EX}{\mathbb{E}}
\newcommand{\appendixref}[1]{\hyperref[#1]{Appendix \ref*{#1}}}
\newcommand{\frechet}{Fr\`echet Inception Distance }

\def\GCPR20SubNumber{71}

\newcommand{\paperTitle}{Image Inpainting with Learnable Feature Imputation}
\newcommand{\paperAuthor}{H\aa kon Hukkel\aa s \orcidID{0000-0001-9830-4931} \and Frank Lindseth \orcidID{0000-0002-4979-9218} \and Rudolf Mester \orcidID{0000-0002-6932-0606}}
\newcommand{\paperInstitute}{
	Department of Computer Science\\
	Norwegian University of Science and Technology \\
	Trondheim, Norway\\
	\email{\{hakon.hukkelas, rudolf.mester, frankl\}@ntnu.no}}
\newcommand{\runningTitle}{Image Inpainting with Learnable Feature Imputation}
\newcommand{\runningAuthor}{H. Hukkel\aa s \and F. Lindseth \and R. Mester}

\begin{document}
\pagestyle{headings}
\mainmatter


\title{\paperTitle}

\titlerunning{\runningTitle}
\authorrunning{\runningAuthor}
\author{\paperAuthor}
\institute{\paperInstitute}

\maketitle

\begin{figure}
\centering
\includegraphics[width=.715\textwidth]{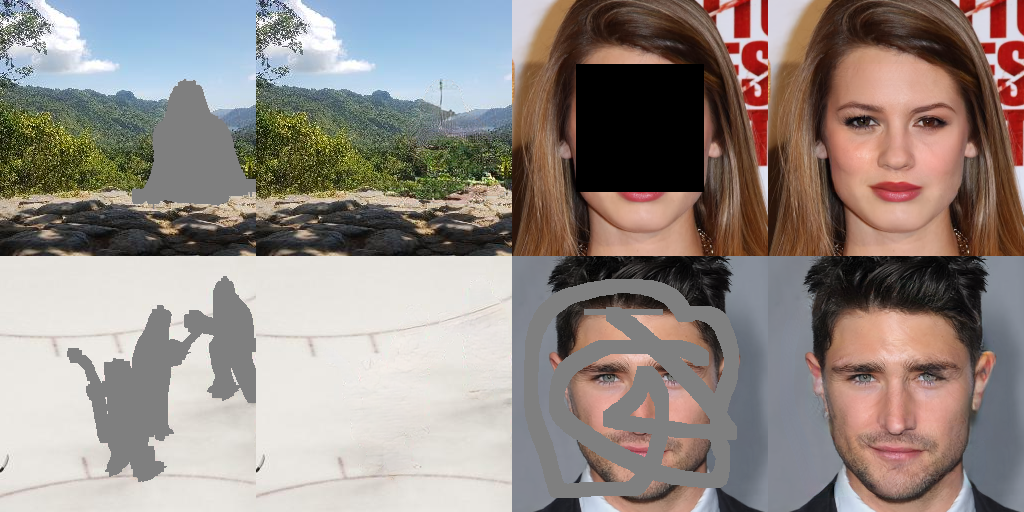}
\caption{Masked images and corresponding generated images from our proposed single-stage generator.}
\label{fig:showoff}
\end{figure}
\begin{abstract}
A regular convolution layer applying a filter in the same way over known and unknown areas causes visual artifacts in the inpainted image.
Several studies address this issue with feature re-normalization on the output of the convolution.
However, these models use a significant amount of learnable parameters for feature re-normalization \cite{BidirectionalAttentionXie,GatedConvolutionYu}, or assume a binary representation of the certainty of an output \cite{Guo_2019,PconvLiu}.

We propose (layer-wise) feature imputation of the missing input values to a convolution.
In contrast to learned feature re-normalization \cite{BidirectionalAttentionXie,GatedConvolutionYu}, our method is efficient and introduces a minimal number of parameters.
Furthermore, we propose a revised gradient penalty for image inpainting, and a novel GAN architecture trained exclusively on adversarial loss.
Our quantitative evaluation on the FDF dataset reflects that our revised gradient penalty and alternative convolution improves generated image quality significantly.
We present comparisons on CelebA-HQ and Places2 to current state-of-the-art to validate our model.
\footnote{
	Code is available at: \href{https://github.com/hukkelas/DeepPrivacy}{github.com/hukkelas/DeepPrivacy}.
	Supplementary material can be downloaded from: \href{https://folk.ntnu.no/haakohu/GCPR\_supplementary.pdf}{folk.ntnu.no/haakohu/GCPR\_supplementary.pdf}}

\end{abstract}
\ifbool{draft}{compiled on \today, \currenttime}{}
\section{Introduction}

Image inpainting is the task of filling in missing areas of an image.
Use cases for image inpainting are diverse, such as restoring damaged images, removing unwanted objects, or replacing information to preserve the privacy of individuals.
Prior to deep learning, image inpainting techniques were generally examplar-based.
For example, pattern matching, by searching and replacing with similar patches \cite{Barnes_2009,Efros_2001,Kwatra_2005,Le_Meur_2011,Simakov_2008,Zongben_Xu_2010}, or diffusion-based, by smoothly propagating information from the boundary of the missing area \cite{Ballester_2001,bertalmio2000image,Criminisi_2004}.

Convolutional Neural Networks (CNNs) for image inpainting have led to significant progress in the last couple of years  \cite{Habibi_Aghdam_2017,K_hler_2014,xie2012image}.
In spite of this, a standard convolution does not consider if an input pixel is missing or not, making it ill-fitted for the task of image inpainting.
Partial Convolution (PConv) \cite{PconvLiu} propose a modified convolution, where they zero-out invalid (missing) input pixels and re-normalizes the output feature map depending on the number of valid pixels in the receptive field.
This is followed by a hand-crafted certainty propagation step, where they assume an output is valid if one or more features in the receptive field are valid.
Several proposed improvements replace the hand-crafted components in PConv with fully-learned components \cite{BidirectionalAttentionXie,GatedConvolutionYu}.
However, these solutions use $\sim 50\%$ of the network parameters to propagate the certainties through the network.

\begin{figure}[t]
\centering
\begin{subfigure}[t]{0.3\textwidth}
\includegraphics[width=\textwidth]{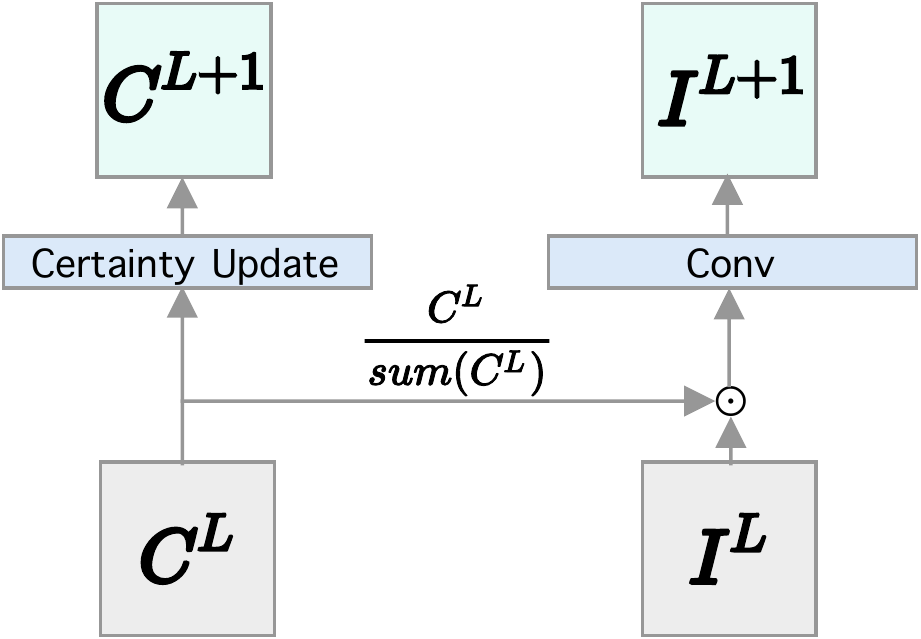}
\caption{Pconv \cite{PconvLiu}}
\end{subfigure}
\hspace{.01\textwidth}
\vrule
\hspace{.01\textwidth}
\begin{subfigure}[t]{0.3\textwidth}
\includegraphics[width=\textwidth]{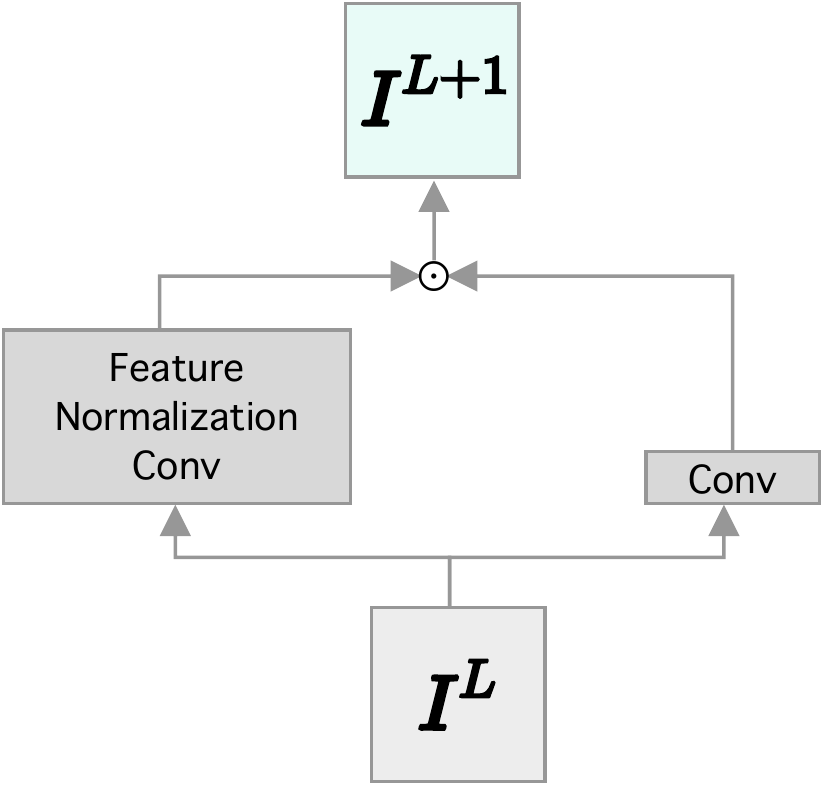}
\caption{Gated Conv \cite{GatedConvolutionYu}}
\end{subfigure}
\hspace{.01\textwidth}
\vrule
\hspace{.01\textwidth}
\begin{subfigure}[t]{0.3\textwidth}
	\includegraphics[width=\textwidth]{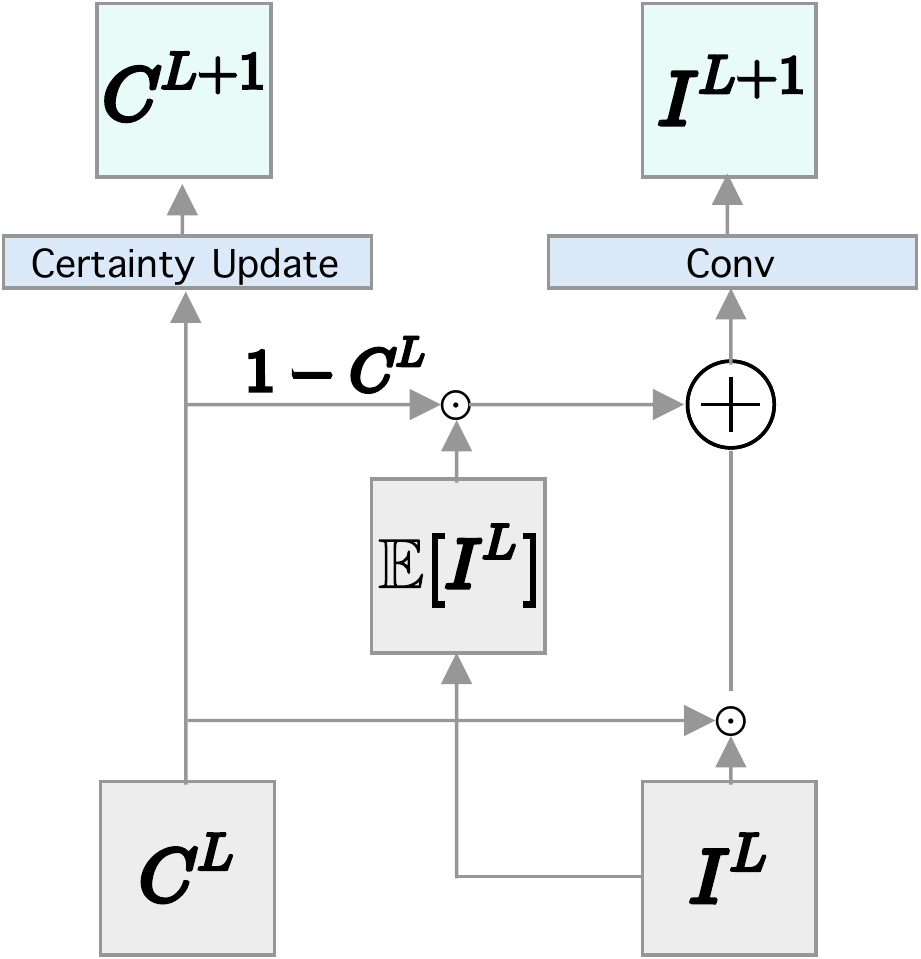}
	\caption{Ours}
\end{subfigure}
\caption{Illustration of partial convolution, gated convolution and our proposed solution.
	$\odot \;$ is element-wise product and $\oplus$ is addition.
Note that $C^L \;$ is binary for partial convolution.}
\label{fig:method_comparison}
\end{figure}

We propose \emph{Imputed Convolution (IConv)}; instead of re-normalizing the output feature map of a convolution, we replace uncertain input values with an estimate from spatially close features (see \autoref{fig:method_comparison}).
IConv assumes that a single spatial location (with multiple features) is associated with a single certainty.
In contrast, previous solutions \cite{BidirectionalAttentionXie,GatedConvolutionYu} requires a certainty \emph{for each feature} in a spatial location, which allocates half of the network parameters for certainty representation and propagation.
Our simple assumption enables certainty representation and propagation to be minimal.
In total, replacing all convolution layers with IConv increases the number of parameters by only $1-2\%$.

We use the DeepPrivacy \cite{hukkelaas2019DeepPrivacy} face inpainter as our baseline and suggest several improvements to stabilize the adversarial training:
(1) We propose an improved version of gradient penalties to optimize Wasserstein GANs \cite{arjovsky2017wasserstein}, based on the simple observation that standard gradient penalties causes training instability for image inpainting.
(2) We combine the U-Net \cite{ronneberger2015u} generator with Multi-Scale-Gradient GAN (MSG-GAN) \cite{karnewar2019msg} to enable the
discriminator to attend to multiple resolutions simultaneously, ensuring global and local consistency.
(3) Finally, we replace the inefficient representation  of the pose-information for the FDF dataset \cite{hukkelaas2019DeepPrivacy}.
In contrast to the current state-of-the-art, our model requires no post-processing of generated images \cite{Iizuka_2017,FaceCompletionLi}, no refinement network \cite{ConextualAttention2018Yu,GatedConvolutionYu}, or any additional loss term to stabilize the adversarial training \cite{BidirectionalAttentionXie,GatedConvolutionYu}.
From our knowledge, our model is the first to be trained exclusively on adversarial loss for image-inpainting.

Our main contributions are the following:
\begin{enumerate}
	\item We propose IConv which utilize a learnable feature estimator to impute uncertain input values to a convolution.
	This enables our model to generate visually pleasing images for free-form image inpainting.
	
	\item We revisit the standard gradient penalty used to constrain Wasserstein GANs for image inpainting.
	Our simple modification significantly improves training stability and generated image quality at no additional computational cost.
	
	\item We propose an improved U-Net architecture, enabling the adversarial training to attend to local and global consistency simultaneously.
\end{enumerate}

\section{Related Work}
In this section, we discuss related work for generative adversarial networks (GANs), GAN-based image-inpainting, and the recent progress in free-form image-inpainting.


\subsubsection{Generative Adversarial Networks}
Generative Adversarial Networks \cite{goodfellow2014generative} is a successful unsupervised training technique for image-based generative models.
Since its conception, a range of techniques has improved convergence of GANs.
Karras \etal \cite{karras2018progressive} propose a \emph{progressive growing} training technique to iteratively increase the network complexity to stabilize training. 
Karnewar \etal \cite{karnewar2019msg} replace progressive growing with  Multi-Scale Gradient GAN (MSG-GAN), where they use skip connections between the matching resolutions of the generator and discriminator.
Furthermore, Karras \etal \cite{karras2019analyzing} propose a modification of MSG-GAN in combination with residual connections \cite{He_2016}.
Similar to \cite{karras2019analyzing}, we replace progressive growing in the baseline model \cite{hukkelaas2019DeepPrivacy} with a modification of MSG-GAN for image-inpainting.

\subsubsection{GAN-based Image Inpainting}
GANs have seen wide adaptation for the image inpainting task, due to its astonishing ability to generate semantically coherent results for missing regions.
There exist several studies proposing methods to ensure global and local consistency;
using several discriminators to focus on different scales \cite{Iizuka_2017,FaceCompletionLi},
specific modules to connect spatially distant features \cite{Song_2018,Yan_2018,Yang_2017,ConextualAttention2018Yu},
patch-based discriminators \cite{GatedConvolutionYu,Zeng_2019},
multi-column generators \cite{wang2018image},
or progressively inpainting the missing area \cite{Guo_2019,Zhang_2018}.
In contrast to these methods, we ensure consistency over multiple resolutions by connecting different resolutions of the generator with the discriminator.
Zheng \etal \cite{zheng2019pluralistic} proposes a probabilistic framework to address the issue of mode collapse for image inpainting, and they generate several plausible results for a missing area.
Several methods propose combining the input image with auxiliary information, such as user sketches \cite{SC-FEGANJo}, edges \cite{nazeri2019edgeconnect}, or examplar-based inpainting \cite{Dolhansky_2018}.
Hukkel\aa s \etal \cite{hukkelaas2019DeepPrivacy} propose a U-Net based generator conditioned on the pose of the face.

GANs are notoriously difficult to optimize reliably \cite{salimans2016improved}.
For image inpainting, the adversarial loss is often combined with other objectives to improve training stability, such as pixel-wise reconstruction \cite{Dolhansky_2018,Iizuka_2017,FaceCompletionLi,Pathak_2016},
 perceptual loss \cite{Song_2018,zhang2018perceptual},
semantic loss \cite{FaceCompletionLi},
or style loss \cite{BidirectionalAttentionXie}.
In contrast to these methods, we optimize exclusively on the adversarial loss.
Furthermore, several studies \cite{SC-FEGANJo,wang2018image,BidirectionalAttentionXie,ConextualAttention2018Yu} propose to use Wasserstein GAN \cite{arjovsky2017wasserstein}  with gradient penalties \cite{gulrajani2017improved}; however, the standard gradient penalty causes training instability for image-inpainting models, as we discuss in Section \ref{sec:gradient_penalty}.

\subsubsection{Free-Form Image-Inpainting}
Image Inpainting with irregular masks (often referred to as free-form masks) has recently caught more attention.
Liu \etal \cite{PconvLiu} propose Partial Convolutions (PConv) to handle irregular masks, where they zero-out input values to a convolution and then perform feature re-normalization based on the number of valid pixels in the receptive field.
Gated Convolution \cite{GatedConvolutionYu} modifies PConv by removing the binary-representation constraint, and they combine the mask and feature representation within a single feature map.
Xie \etal \cite{BidirectionalAttentionXie} propose a simple modification to PConv, where they reformulate it as "attention" propagation instead of certainty propagation.
Both of these PConv adaptations \cite{BidirectionalAttentionXie,GatedConvolutionYu} doubles the number of parameters in the network when replacing regular convolutions.

\section{Method}
In this section, we describe a) our modifications to a regular convolution layer, b) our revised gradient penalty suited for image inpainting, and c) our improved U-Net architecture.

\subsection{Imputed Convolution (IConv)}

Consider the case of a regular convolution applied to a given feature map $I \in \mathbb{R}^{N}$:
\begin{equation}
f(I) = W_F * I,
\end{equation}
where $* \;$ is the convolution and $W_F \in \mathbb{R}^{D}$ is the filter.
To simplify notation, we consider a single filter applied to a single one-dimensional feature map.
The generalization to a regular multidimensional convolution layer is straightforward.
A convolution applies this filter to all spatial locations of our feature map, which works well for general image recognition tasks.
For image inpainting, there exists a set of known and unknown pixels;
therefore, a regular convolution applied to all spatial locations is primarily undefined (``unknown" is not the same as 0 or any other fixed value), and naive approaches cause annoying visual artifacts \cite{PconvLiu}.

We propose to replace the missing input values to a convolution with an estimate from spatially close values.
To represent known and unknown values, we introduce a certainty $C_x$ for each spatial location $x$, where $C \in \mathbb{R}^N$, and $0 \leq C_x \leq 1$.
Note that this representation enables a single certainty to represent several values in the case of having multiple channels in the input.
Furthermore, we define $\tilde{I}_x$ as a random variable with discrete outcomes $\{I_x, h_x\}$, where $I_x$ is the feature at spatial location $x$, and $h_x$ is an estimate from spatially close features.
In this way, we want the output of our convolution to be given by,
\begin{equation}
O = \phi (f(\EX[\tilde{I_x}])),
\label{eq:our_method}
\end{equation}
where $\phi$ is the activation function, and $O$ the output feature map.
We approximate the probabilities of each outcome using the certainty $C_x$; that is, $P(\tilde{I_x} = I_x) \approx C_x$ and $P(\tilde{I_x} = h_x) \approx 1 - C_x$, yielding 
the expected value of $\tilde{I}_x$,
\begin{equation}
\EX[\tilde{I_x}] ~= C_x \cdot I_x + (1-C_x) \cdot h_x.
\label{eq:linear_interpolation}
\end{equation}
We assume that a missing value can be approximated from spatially close values.
Therefore, we define $h_x$ as a learned certainty-weighted average of the surrounding features:
\begin{equation}
h_x = \frac{
	\sum_{i=1}^{K} I_{x+i} \cdot C_{x+i} \cdot \omega_i
}{
	\sum_{i=1}^K  C_{x+i}
},
\label{eq:h_i}
\end{equation} 
where $\omega \in R^{K}$ is a learnable parameter.
In a sense, our convolutional layer will try to learn the outcome space of $\tilde{I_x}$. 
Furthermore, $h_x$ is efficient to implement in standard deep learning frameworks, as it can be implemented as a depth-wise separable convolution \cite{DeptwiseConvSifre} with a re-normalization factor determined by $C$.

\subsubsection{Propagating Certainties}

Each convolutional layer expects a certainty for each spatial location.
We handle propagation of certainties as a learned operation,
\begin{equation}
C^{L+1} = \sigma (W_C * C^{L}),
\label{eq:certainty_propagation}
\end{equation}
where $*$ is a convolution, $W_C \in \mathbb{R}^{D}$ is the filter, and $\sigma$ is the sigmoid function.
We constraint $W_C$ to have the same receptive field as $f$ with no bias, and initialize $C^0\;$ to 0 for all unknown pixels and 1 else.

The proposed solution is minimal, efficient, and other components of the network remain close to untouched.
We use LeakyReLU as the activation function $\phi$, and average pooling and pixel normalization \cite{karras2018progressive} after each convolution $f$.
Replacing all convolutional layers with $O_x$ in our baseline network increases the number of parameters by $\sim 1\%$.
This is in contrast to methods based on learned feature re-normalization \cite{BidirectionalAttentionXie,GatedConvolutionYu}, where replacing a convolution with their proposed solution doubles the number of parameters.
Similar to partial convolution \cite{PconvLiu}, we use a single scalar to represent the certainty for each spatial location; however, we do not constrain the certainty representation to be binary, and our certainty propagation is fully learned.

\subsubsection{U-Net Skip Connection}
U-Net \cite{ronneberger2015u} skip  connection is a method to combine shallow and deep features in encoder-decoder architectures.
Generally, the skip connection consists of concatenating shallow and deep features, then followed by a convolution.
However, for image inpainting, we only want to propagate certain features.

To find the combined feature map for an input in layer $L$ and $L+l$, we find a weighted average.
Assuming features from two layers in the network, $(I^L, C^L)$, $(I^{L+l}, C^{L+l})$, we define the combined feature map as;
\begin{equation}
I^{L+l+1} = \gamma \cdot I^L + (1-\gamma) \cdot I^{L+l},
\label{eq:unet_skip}
\end{equation}
and likewise for $C^{L+l+1}$. 
$\; \gamma \;$ is determined by
\begin{equation}
\gamma = \frac{C^L \cdot \beta_1}{C^L \cdot \beta_1 + C^{L+l} \cdot \beta_2},
\end{equation}
where $\beta_1, \beta_2 \in \mathbb{R^+}$ are learnable parameters initialized to 1.
Our U-Net skip connection is unique compared to previous work and designed for image inpainting.
\autoref{eq:unet_skip} enables the network to only propagate features with a high certainty from shallow layers.
Furthermore, we  include $\beta_1$ and $\beta_2\; $ to give the model the flexibility to learn if it should attend to shallow or deep features.

\subsection{Revisiting Gradient Penalties for Image Inpainting}
\label{sec:gradient_penalty}
Improved Wasserstein GAN \cite{arjovsky2017wasserstein,gulrajani2017improved} is widely used in image inpainting \cite{SC-FEGANJo,wang2018image,BidirectionalAttentionXie,ConextualAttention2018Yu}.
Given a discriminator $D$, the objective function for optimizing a Wasserstein GAN with gradient penalties is given by,
\begin{equation}
	\mathcal{L}_{total} = \mathcal{L}_{adv} +
	\lambda \cdot (||\nabla D(\hat{x})||_p  - 1)^2,
\end{equation}
where $\mathcal{L}_{adv}$ is the adversarial loss, $p$ is commonly  set to 2 
($L^2$ norm), 
$\lambda$ is the gradient penalty weight, and $\hat{x}$ is a randomly sampled point between the real image, $x$,$\;$ and a generated image, $\tilde{x}$.
Specifically, $\hat{x} = t \cdot x + (1-t)\cdot \tilde{x}$, where $t$ is sampled from a uniform distribution \cite{gulrajani2017improved}.

Previous methods enforce the gradient penalty only for missing areas \cite{SC-FEGANJo,wang2018image,ConextualAttention2018Yu}.
Given a mask $M$ to indicate areas to be inpainted in the image $x$, where $M \;$ is 0 for missing pixels and 1 otherwise (note that $M=C^0$), Yu \etal \cite{ConextualAttention2018Yu} propose the gradient penalty:
\begin{equation}
\bar{g}(\hat{x}) = (||\nabla D(\hat{x})\odot (1-M)||_p  - 1)^2,
\label{eq:gp_standard}
\end{equation}
where $\odot$ is element-wise multiplication.
This gradient penalty cause significant training instability, as the gradient sign of $\bar{g}$ shifts depending on the cardinality of $M$.
Furthermore, \autoref{eq:gp_standard} impose $||\nabla D(\hat{x})|| \approx 1$, which leads to a lower bound on the Wasserstein distance \cite{jolicoeur2019connections}.

Imposing $||\nabla D(\hat{x})|| \leq 1$ will remove the issue of shifting gradients in \autoref{eq:gp_standard}.
Furthermore, imposing the constrain  $||\nabla D(\hat{x})|| \leq 1$ is shown to properly estimate the Wasserstein distance \cite{jolicoeur2019connections}.
Therefore, we propose the following gradient penalty:
\begin{equation}
	g(\hat{x}) = \max(0, ||\nabla D(\hat{x})\odot (1-M))||_p -1)
	\label{eq:linf_clamp}
\end{equation}
Previous methods enforce the $L^2$ norm \cite{SC-FEGANJo,wang2018image,ConextualAttention2018Yu}.
Jolicoeur-Martineau \etal \cite{jolicoeur2019connections} suggest that replacing the $L^2$ gradient norm with $L^\infty$ can improve robustness.
From empirical experiments (see Appendix 1), we find $L^\infty$ more unstable and sensitive to choice of hyperparameters; therefore, we enforce the $L^2$ norm (p=2).

In total, we optimize the following objective function:
\begin{equation}
\mathcal{L}_{total} = \mathcal{L}_{adv} +
\lambda \cdot \max(0, ||\nabla D(\hat{x})\odot (1-M))||_p -1)
\end{equation}

\begin{figure}[t]
\centering
\includegraphics[width=\textwidth]{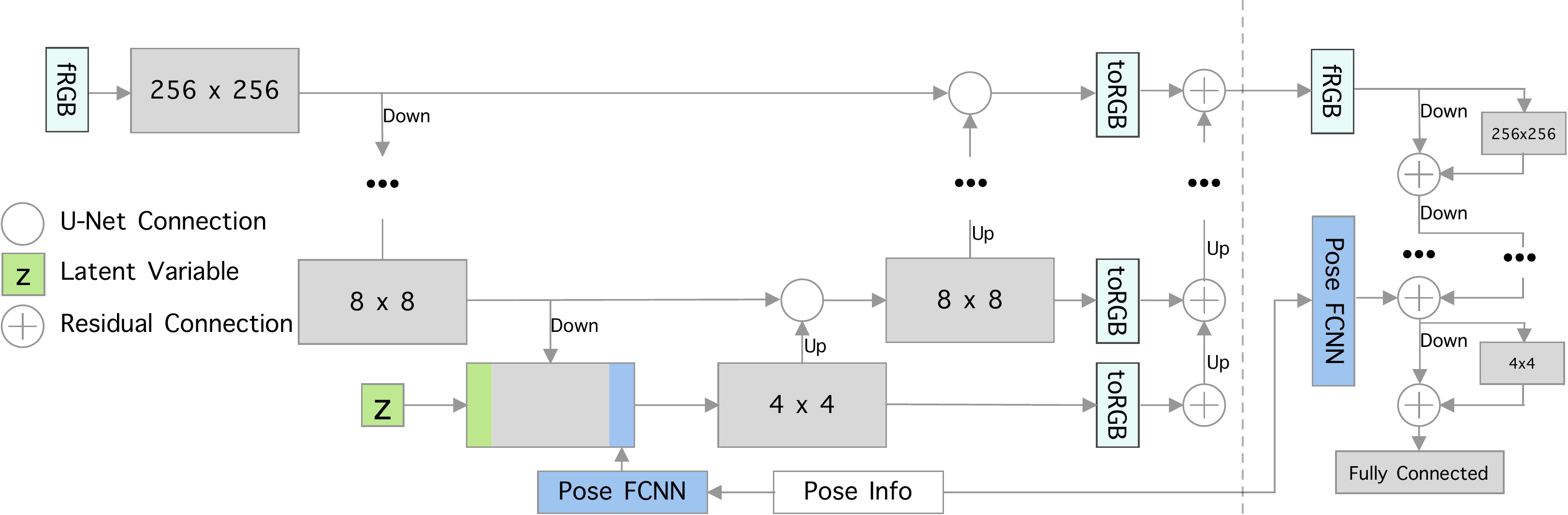}
\caption{
Illustration of the generator (left of the dashed line) and discriminator architecture.
Up and down denotes nearest neighbor upsampling and average pool.
The pose information in the discriminator is concatenated to the input of the first convolution layer with $32 \times 32 \;$ resolution.
Note that pose information is only used for the FDF dataset \cite{hukkelaas2019DeepPrivacy}.
}
\label{fig:model_architecture}
\end{figure}
\subsection{Model Architecture}

We propose several improvements to the baseline U-Net architecture \cite{hukkelaas2019DeepPrivacy}.
See \autoref{fig:model_architecture} for our final architecture.
We replace all convolutions with \autoref{eq:our_method}, average pool layer with a certainty-weighted average and U-Net skip connections with our revised skip connection (see \autoref{eq:unet_skip}).
Furthermore, we replace progressive growing training \cite{karras2018progressive} with Multi-Scale Gradient GAN (MSG-GAN) \cite{karnewar2019msg}.
For the MSG-GAN, instead of matching different resolutions from the generator with the discriminator, we upsample each resolution and sum up the contribution of the RGB outputs \cite{karras2019analyzing}.
In the discriminator we use residual connections, similar to \cite{karras2019analyzing}.
Finally, we improve the representation of pose information in the baseline model (pose information is only used on the FDF dataset \cite{hukkelaas2019DeepPrivacy}).

\subsubsection{Representation of Pose Information}
The baseline model \cite{hukkelaas2019DeepPrivacy} represents pose information as one-hot encoded images for each resolution in the network, which is extremely memory inefficient and a fragile representation.
The pose information, $P \in \mathbb{R}^{K\cdot2}$, represents K facial keypoints and is used as conditional information for the generator and discriminator.
We propose to replace the one-hot encoded representation, and instead pre-process $P$ into a $4 \times 4 \times 32 \;$  feature bank using two fully-connected layers.
This feature bank is concatenated with the features from the encoder.
Furthermore, after replacing progressive growing with MSG-GAN, we include the same pose pre-processing architecture in the discriminator, and input the pose information as a $32 \times 32 \times 1$ feature map to the discriminator.

\section{Experiments}

We evaluate our proposed improvements on the Flickr Diverse Faces (FDF) dataset \cite{hukkelaas2019DeepPrivacy}, a lower resolution ($128 \times 128$) face dataset.
We present experiments on the CelebA-HQ \cite{karras2018progressive} and Places2 \cite{places2} datasets, which reflects that our suggestions generalizes to standard image inpainting.
We compare against current state-of-the art \cite{BidirectionalAttentionXie,GatedConvolutionYu,zheng2019pluralistic,Ren_2019}.
Finally, we present a set of ablation studies to analyze the generator architecture.
\footnote{
	To prevent ourselves from cherry-picking qualitative examples, we present several images (with corresponding masks) chosen by previous state-of-the-art papers
	\cite{Guo_2019,BidirectionalAttentionXie,GatedConvolutionYu,zheng2019pluralistic},
	thus copying their selection.
	Appendix 5 describes how we selected these samples.
	The only hand-picked examples in this paper are \autoref{fig:showoff}, \autoref{fig:cherry_0}, \autoref{fig:deterministc_experiment}, and \autoref{fig:unet_skip_connections}.
	No examples in the Supplementary Material are cherry-picked.
}

\subsubsection{Quantitative Metrics}
For quantitative evaluations, we report commonly used image inpainting metrics; pixel-wise distance (L1 and L2), peak signal-to-noise ratio (PSNR), and structural similarity (SSIM).
Neither of these reconstruction metrics are any good indicators of generated image quality, as there often exist several possible solutions to a missing region, and they do not reflect human nuances \cite{zhang2018perceptual}.
Recently proposed deep feature metrics correlate better with human perception \cite{zhang2018perceptual}; therefore, we report the  \frechet (FID) \cite{FID2017heusel} (lower is better) and Learned Perceptual Image Patch Similarity (LPIPS) \cite{zhang2018perceptual} (lower is better).
We use LPIPS as the main quantitative evaluation.

\begin{table}[t]
	
	\caption{
		\textbf{Quantitative results on the FDF dataset} \cite{hukkelaas2019DeepPrivacy}.
		We report standard metrics after showing the discriminator 20 million images on the FDF and Places2 validation sets.
		We report L1, L2, and SSIM in Appendix 3.
		Note that Config E is trained with MSG-GAN, therefore, we separate it from Config A-D which are trained with progressive growing \cite{karras2018progressive}.
		* Did not converge.
		$\dagger$ Same as Config B}
	\centering
\begin{tabular}{ll|ccc|ccc}
	\hline
	\multicolumn{2}{l|}{Configuration} & \multicolumn{3}{c|}{FDF } & \multicolumn{3}{c}{Places2} \\ 
	\hline
	&  & LPIPS $\downarrow$ & PSNR $\uparrow$ & FID $\downarrow$ & LPIPS $\downarrow$ & PSNR $\uparrow$ & FID $\downarrow$ \\
	A &  Baseline \cite{hukkelaas2019DeepPrivacy} & 0.1036 & 22.52 & 6.15 & --* & --* & --* \\
	B &  + Improved Gradient penalty & 0.0757 & 23.92 & 1.83 & 0.1619 & 20.99 & 7.96 \\
	C & + Scalar Pose Information & 0.0733 & 24.01 & 1.76 & -- $\dagger$ & --$\dagger$ & -- $\dagger$ \\
	D & + Imputed Convolution & 0.0739 & 23.95 & 1.66 & 0.1563 & 21.21 & 6.81 \\
	\hline
	E & + No Growing, MSG & \textbf{0.0728} & \textbf{24.01} & \textbf{1.49} & \textbf{0.1491} & \textbf{21.42} & \textbf{5.24} \\

\end{tabular}
\label{tab:configurations}
\end{table}

\subsection{Improving the Baseline}
We iteratively add our suggestions to the baseline \cite{hukkelaas2019DeepPrivacy} (Config A-E), and report quantitative results in \autoref{tab:configurations}.
First, we replace the gradient penalty term with \autoref{eq:linf_clamp}, where we use the $L^2\;$ norm ($p=2$), and impose the following constraint (Config B):
\begin{equation}
	G_{out} = G(I, C^0) \cdot (1-C^0) + I \cdot C^0,
\end{equation}
where $C^0$ is the binary input certainty and $G$ is the generator.
Note that we are not able to converge Config A while imposing $G_{out}$.
We replace the one-hot encoded representation of the pose information with two fully connected layers in the generator (Config C).
Furthermore, we replace the input to all convolutional layers with \autoref{eq:linear_interpolation} (Config D).
We set the receptive field of $h_x$ to $5 \times 5 \;$ ($K=5$ in \autoref{eq:h_i}).
We replace the progressive-growing training technique with MSG-GAN \cite{karnewar2019msg}, and replace the one-hot encoded pose-information in the discriminator (Config E).
These modifications combined improve the LPIPS score by \emph{30.0\%}.
The authors of \cite{hukkelaas2019DeepPrivacy} report a FID of $1.84 \;$ on the FDF dataset $\;$ with a model consisting of 46M learnable parameters.
In comparison, we achieve a FID of $1.49 \;$ with 2.94M parameters (config E).
For experimental details, see Appendix 2.

\begin{figure}[!ht]
\centering
\begin{subfigure}[t]{0.2\textwidth}
\includegraphics[width=\textwidth]{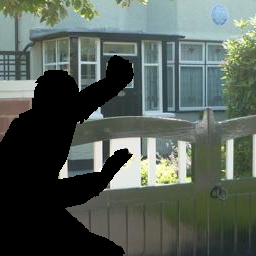}
\includegraphics[width=\textwidth]{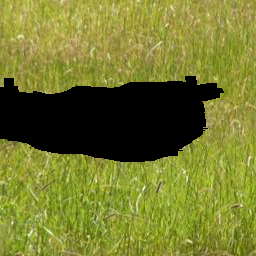}
\includegraphics[width=\textwidth]{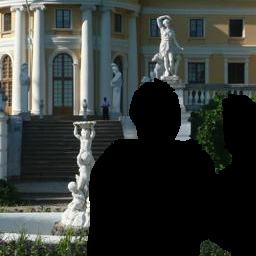}
\includegraphics[width=\textwidth]{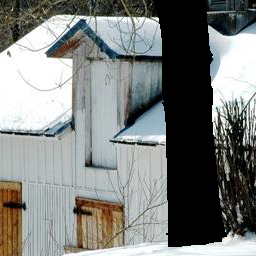}
\includegraphics[width=\textwidth]{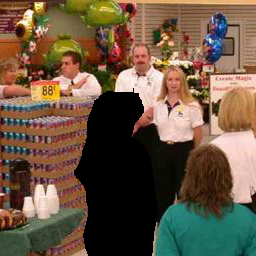}
\includegraphics[width=\textwidth]{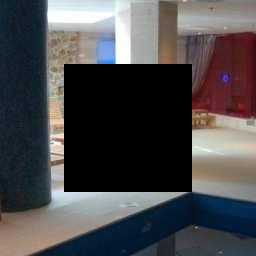}
\caption{Input}
\end{subfigure}
\hspace{-0.005\textwidth}
\begin{subfigure}[t]{0.2\textwidth}
\includegraphics[width=\textwidth]{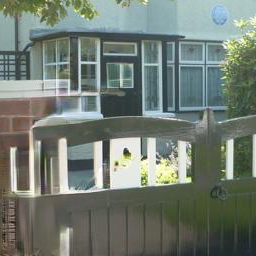}
\includegraphics[width=\textwidth]{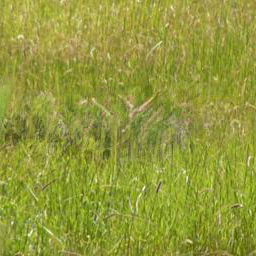}
\includegraphics[width=\textwidth]{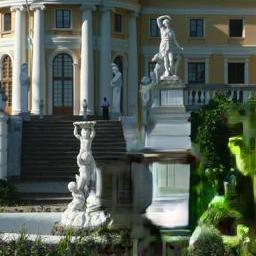}
\includegraphics[width=\textwidth]{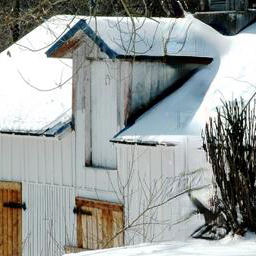}
\includegraphics[width=\textwidth]{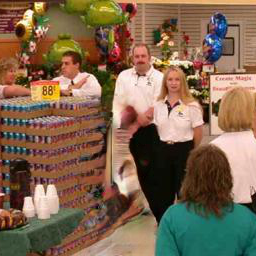}
\includegraphics[width=\textwidth]{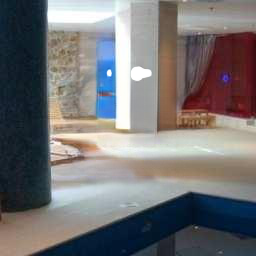}
\caption{GConv \cite{GatedConvolutionYu}}
\end{subfigure}
\hspace{-0.005\textwidth}
\begin{subfigure}[t]{0.2\textwidth}
\includegraphics[width=\textwidth]{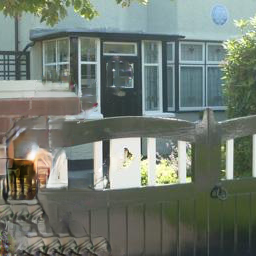}
\includegraphics[width=\textwidth]{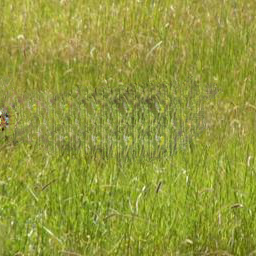}
\includegraphics[width=\textwidth]{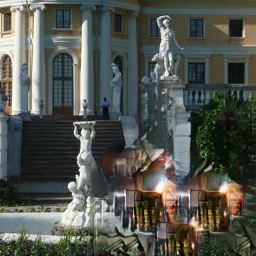}
\includegraphics[width=\textwidth]{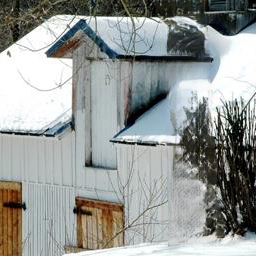}
\includegraphics[width=\textwidth]{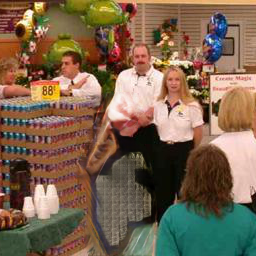}
\includegraphics[width=\textwidth]{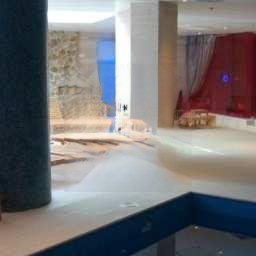}
\caption{PIC \cite{zheng2019pluralistic}}
\end{subfigure}
\hspace{-0.005\textwidth}
\begin{subfigure}[t]{0.2\textwidth}
\includegraphics[width=\textwidth]{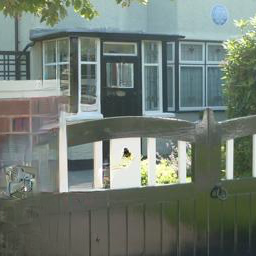}
\includegraphics[width=\textwidth]{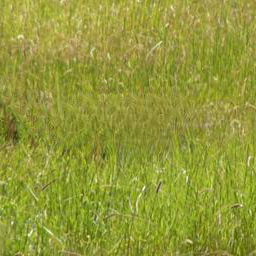}
\includegraphics[width=\textwidth]{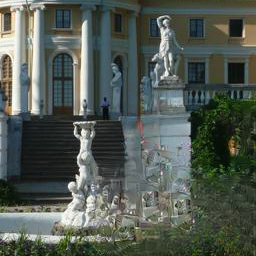}
\includegraphics[width=\textwidth]{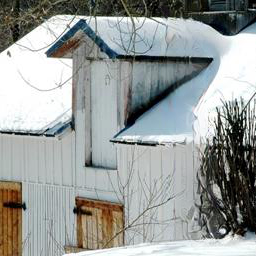}
\includegraphics[width=\textwidth]{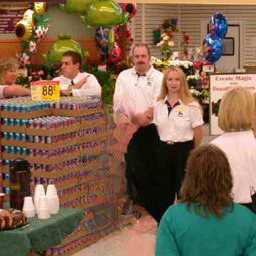}
\includegraphics[width=\textwidth]{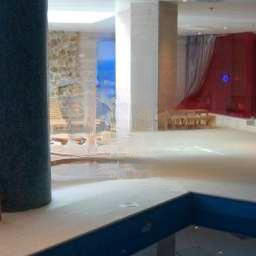}
\caption{SF \cite{Ren_2019}}
\end{subfigure}
\hspace{-0.005\textwidth}
\begin{subfigure}[t]{0.2\textwidth}
\includegraphics[width=\textwidth]{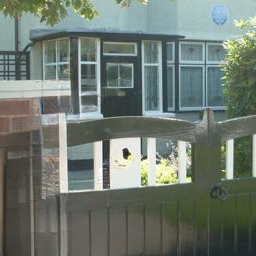}
\includegraphics[width=\textwidth]{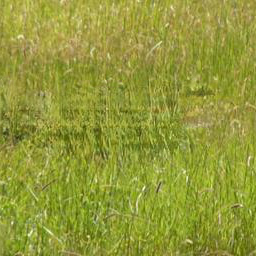}
\includegraphics[width=\textwidth]{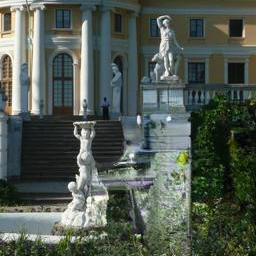}
\includegraphics[width=\textwidth]{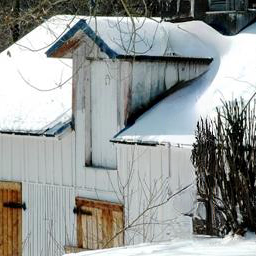}
\includegraphics[width=\textwidth]{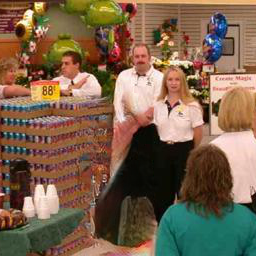}
\includegraphics[width=\textwidth]{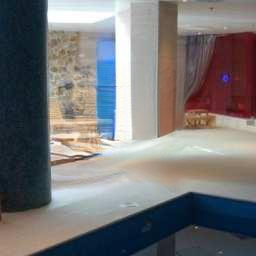}
\caption{Ours}
\end{subfigure}
\hspace{-0.005\textwidth}
\caption{Qualitative examples on the Places2 validation set with comparisons to Gated Convolution (GConv) \cite{GatedConvolutionYu}, StructureFlow (SF) \cite{Ren_2019}, and Pluralistic Image Completion (PIC) \cite{zheng2019pluralistic}.
We recommend the reader to zoom-in on missing regions.
For non hand-picked qualitative examples, see Appendix 5.}
\label{fig:cherry_0}
\end{figure}

\subsection{Generalization to Free-Form Image Inpainting}
We extend Config E to general image inpainting datasets; CelebA-HQ \cite{karras2018progressive} and Places2 \cite{places2}.
We increase the number of filters in each convolution by a factor of 2, such that the generator has $11.5$M $\;$ parameters.
In comparison,
Gated Convolution \cite{GatedConvolutionYu} use $4.1$M,
LBAM \cite{BidirectionalAttentionXie} $68.3$M,
StructureFlow \cite{Ren_2019} $159$M,
and PIC \cite{zheng2019pluralistic}  use $3.6$M parameters.
Compared to \cite{GatedConvolutionYu,zheng2019pluralistic}, our increase in parameters improves semantic reasoning for larger missing regions.
Also, compared to previous solutions, we achieve similar inference time since the majority of the parameters are located at low-resolution layers ($8 \times 8$ and $16 \times 16$).
In contrast, \cite{GatedConvolutionYu} has no parameters at a resolution smaller than $64\times 64$.
For single-image inference time, our model matches (or outperforms) previous models;
on a single NVIDIA 1080 GPU, our network runs at $\; \sim 89 $ ms per image on $256 \times 256$ resolution, $2 \times$ faster than LBAM \cite{BidirectionalAttentionXie}, and PIC \cite{zheng2019pluralistic}.
GatedConvolution \cite{GatedConvolutionYu} achieves $\; \sim 62$ ms per image.
\footnote{We measure runtime for \cite{GatedConvolutionYu,zheng2019pluralistic} with their open-source code, as they do not report inference time for $256 \times 256 \;$ resolution in their paper.}
See Appendix 2.1 for experimental details.

\subsubsection{Quantitative Results}
\autoref{tab:dataset_comparisons} shows quantitative results for the CelebA-HQ and Places2 datasets.
For CelebA-HQ, we improve LPIPS and FID significantly compared to previous models.
For Places2, we achieve comparable results to \cite{GatedConvolutionYu} for free-form and center-crop masks.
Furthermore, we compare our model with and without IConv and notice a significant improvement in generated image quality (see Figure 1 in Appendix 3).
See Appendix 5.1 for examples of the center-crop and free-form images.

\begin{table}[t]
\caption{Quantitative results on the CelebA-HQ and Places2 datasets.
We use the official frameworks to reproduce results from \cite{GatedConvolutionYu,zheng2019pluralistic}.
For the (Center) dataset we use a $128 \times 128$ center mask, and for (Free-Form) we generate free-form masks for each image following the approach in \cite{GatedConvolutionYu}.
We report L1, L2, and SSIM in Appendix 3.}
\centering
\resizebox{\textwidth}{!}{
\begin{tabular}{l|ccc|ccc|ccc|ccc}
\hline
\multirow{2}{*}{Method} & \multicolumn{3}{|c}{Places2 (Center)}  & \multicolumn{3}{|c}{Places2 (Free Form)}  & \multicolumn{3}{|c}{CelebA-HQ (Center)}  & \multicolumn{3}{|c}{CelebA-HQ (Free Form)} \\
 & PSNR & LPIPS & FID & PSNR & LPIPS & FID & PSNR & LPIPS & FID & PSNR & LPIPS & FID\\
\hline
Gated Convolutions \cite{GatedConvolutionYu} & 21.56 & \textbf{0.1407} & 4.14 & \textbf{27.59} & \textbf{0.0579} & 0.90 & \textbf{25.55} & 0.0587 & 6.05 & 30.26 & 0.0366 & 2.98\\
Plurastic Image Inpainting \cite{zheng2019pluralistic} & 21.04 & 0.1584 & 7.23 & 26.66 & 0.0804 & 2.76 & 24.59 & 0.0644 & 7.50 & 29.30 & 0.0394 & 3.30\\
Ours & \textbf{21.70} & 0.1412 & \textbf{3.99} & 27.33 & 0.0597 & 0.94 & 25.29 & \textbf{0.0522} & \textbf{4.43} & \textbf{30.32} & \textbf{0.0300} & \textbf{2.38}\\

\end{tabular}}
\label{tab:dataset_comparisons}
\end{table}

\subsubsection{Qualitative Results}
\begin{figure}[t]
\centering
\begin{subfigure}[t]{0.142\textwidth}
\includegraphics[width=\textwidth]{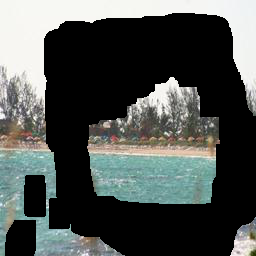}
\includegraphics[width=\textwidth]{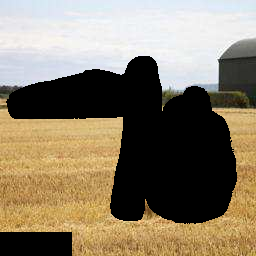}
\caption{Input}
\end{subfigure}
\begin{subfigure}[t]{0.142\textwidth}
\includegraphics[width=\textwidth]{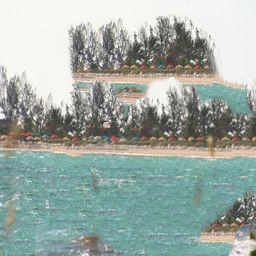}
\includegraphics[width=\textwidth]{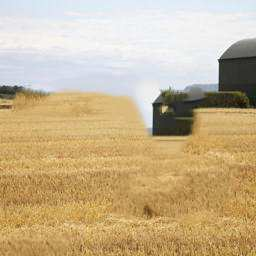}
\caption{PM \cite{Barnes_2009}}
\end{subfigure}
\begin{subfigure}[t]{0.142\textwidth}
\includegraphics[width=\textwidth]{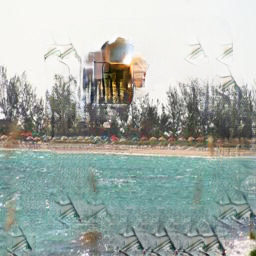}
\includegraphics[width=\textwidth]{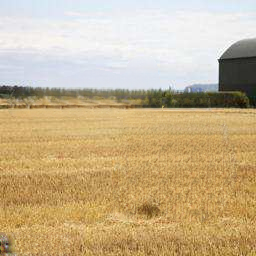}
\caption{PIC \cite{zheng2019pluralistic}}
\end{subfigure}
\begin{subfigure}[t]{0.142\textwidth}
\includegraphics[width=\textwidth]{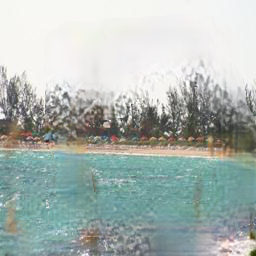}
\includegraphics[width=\textwidth]{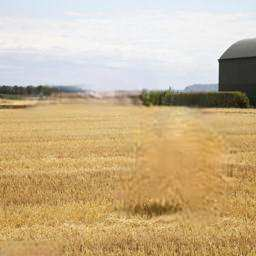}
\caption{PC \cite{PconvLiu}}
\end{subfigure}
\begin{subfigure}[t]{0.142\textwidth}
\includegraphics[width=\textwidth]{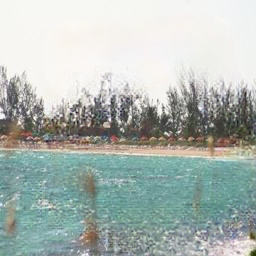}
\includegraphics[width=\textwidth]{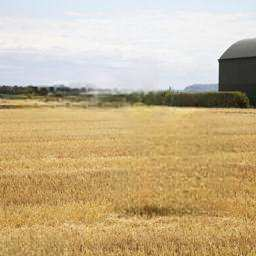}
\caption{BA \cite{BidirectionalAttentionXie}}
\end{subfigure}
\begin{subfigure}[t]{0.142\textwidth}
\includegraphics[width=\textwidth]{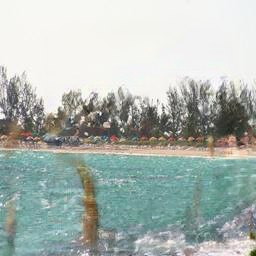}
\includegraphics[width=\textwidth]{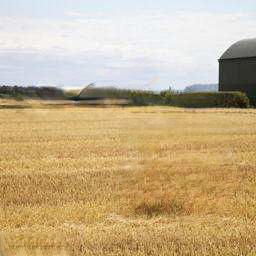}
\caption{GC \cite{GatedConvolutionYu}}
\end{subfigure}
\begin{subfigure}[t]{0.142\textwidth}
\includegraphics[width=\textwidth]{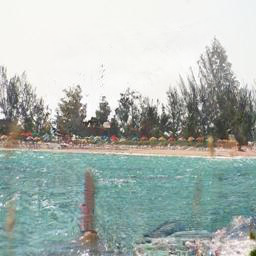}
\includegraphics[width=\textwidth]{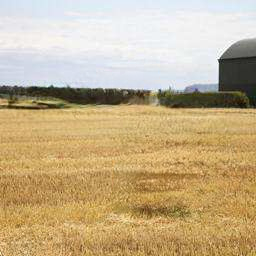}
\caption{Ours}
\end{subfigure}
\caption{Places2 comparison to  PatchMatch (PM) \cite{Barnes_2009}, Pluralistic Image Completion (PIC) \cite{zheng2019pluralistic}, Partial Convolution (PC) \cite{PconvLiu}, Bidirectional Attention (BA) \cite{BidirectionalAttentionXie}, and Gated Convolution (GC) \cite{GatedConvolutionYu}.
Examples selected by authors of \cite{BidirectionalAttentionXie} (images extracted from their supplementary material).
Results of \cite{GatedConvolutionYu,zheng2019pluralistic} generated by using their open-source code and models.
We recommend the reader to zoom-in on missing regions.}
\label{fig:LBAM_all}
\end{figure}

\autoref{fig:cherry_0} shows a set of hand-picked examples, \autoref{fig:LBAM_all} shows examples selected by \cite{BidirectionalAttentionXie}, and  Appendix 5  includes a large set of examples selected by the authors of \cite{Guo_2019,BidirectionalAttentionXie,GatedConvolutionYu,zheng2019pluralistic}.
We notice less visual artifacts than models using vanilla convolutions \cite{zheng2019pluralistic,Ren_2019}, and we achieve comparable results to Gated Convolution \cite{GatedConvolutionYu} for free-form image inpainting.
For larger missing areas, our model generates more semantically coherent results compared to previous solutions \cite{Guo_2019,BidirectionalAttentionXie,GatedConvolutionYu,zheng2019pluralistic}.

\subsection{Ablation Studies}

\subsubsection{Pluralistic Image Inpainting}
\begin{figure}[t]
\centering
\includegraphics[width=\textwidth]{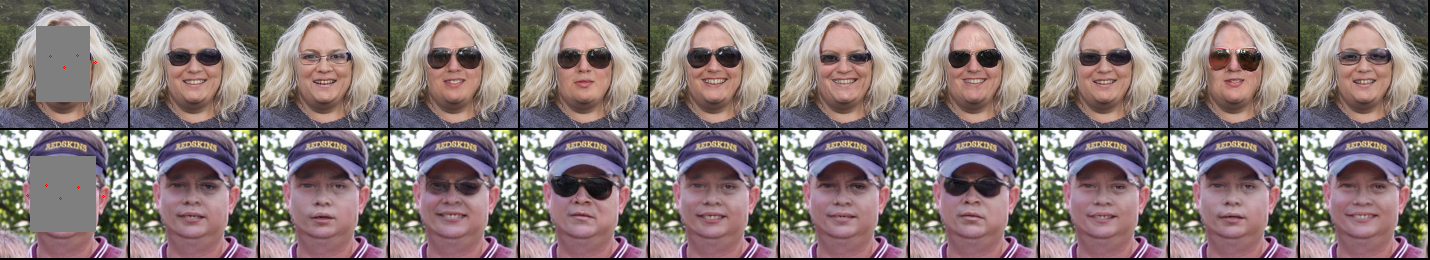}
\caption{
\textbf{Diverse Plausible Results:}
Images from the FDF validation set \cite{hukkelaas2019DeepPrivacy}.
Left column is the input image with the pose information marked in red.
Second column and onwards are different plausible generated results.
Each image is generated by randomly sampling a latent variable for the generator (except for the second column where the latent variable is set to all 0's).
For more results, see  Appendix 6.}
\label{fig:deterministc_experiment}

\end{figure}
Generating different possible results for the same conditional image (pluralistic inpainting) \cite{zheng2019pluralistic} has remained a problem for conditional GANs \cite{huang2018multimodal,zhu2017toward}.
\autoref{fig:deterministc_experiment} illustrates that our proposed model (Config E) generates multiple and diverse results.
Even though, for Places2, we observe that our generator suffers from mode collapse early on in training.
Therefore, we ask the question; \textit{does a deterministic generator impact the generated image quality for image-inpainting?}
To briefly evaluate the impact of this, we train Config D  without a latent variable, and observe a 7\% degradation in LPIPS score on the FDF dataset.
We leave further analysis of this for further work.

\subsubsection{Propagation of Certainties}
\begin{figure}[t]
\centering
\begin{subfigure}[t]{0.138\textwidth}
\includegraphics[width=\textwidth]{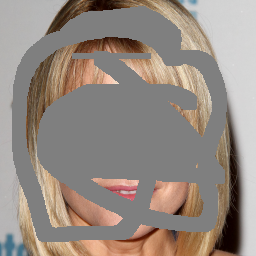}
\end{subfigure}
\begin{subfigure}[t]{0.138\textwidth}
\includegraphics[width=\textwidth]{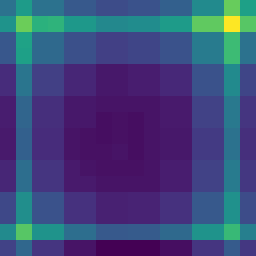}
\end{subfigure}
\begin{subfigure}[t]{0.138\textwidth}
\includegraphics[width=\textwidth]{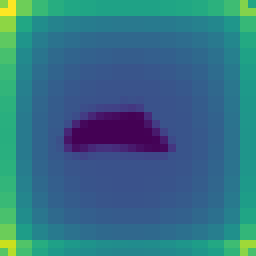}
\end{subfigure}
\begin{subfigure}[t]{0.138\textwidth}
\includegraphics[width=\textwidth]{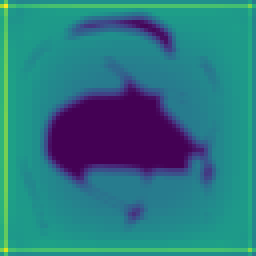}
\end{subfigure}
\begin{subfigure}[t]{0.138\textwidth}
\includegraphics[width=\textwidth]{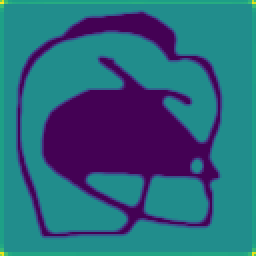}
\end{subfigure}
\begin{subfigure}[t]{0.138\textwidth}
\includegraphics[width=\textwidth]{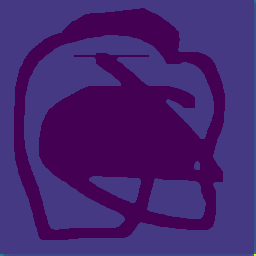}
\end{subfigure}
\begin{subfigure}[t]{0.138\textwidth}
\includegraphics[width=\textwidth]{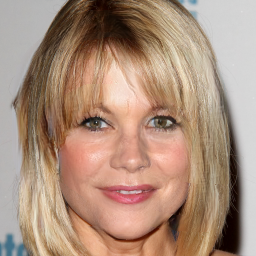}
\end{subfigure}
\begin{subfigure}[t]{0.035\textwidth}
\includegraphics[width=\textwidth]{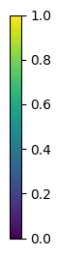}
\end{subfigure}
\caption{
    \textbf{U-Net Skip Connections.}
    Visualization of $\gamma$ from \autoref{eq:unet_skip}.
    The left image is the input image, second column and onwards are the values of $\gamma$ for resolution 8 to 256. Rightmost image is the generated image.
    Smaller values of $\gamma$ indicates that the network selects deep features (from the decoder branch).
}
\label{fig:unet_skip_connections}
\end{figure}

\autoref{fig:unet_skip_connections} visualizes if the generator attends to shallow or deep features in our encoder-decoder architecture.
Our proposed U-Net skip connection enables the network to select features between the encoder and decoder depending on the certainty.
Notice that our network attends to deeper features in cases of uncertain features, and shallower feature otherwise.

\section{Conclusion}
We propose a simple single-stage generator architecture for free-form image inpainting.
Our proposed improvements to GAN-based image inpainting significantly stabilizes adversarial training, and from our knowledge, we are the first to produce state-of-the-art results by exclusively optimizing an adversarial objective.
Our main contributions are; a revised convolution to properly handle missing values in convolutional neural networks, an improved gradient penalty for image inpainting which substantially improves training stability, and a novel U-Net based GAN architecture to ensure global and local consistency.
Our model achieves state-of-the-art results on the CelebA-HQ and Places2 datasets, and our single-stage  generator is much more efficient compared to previous solutions.

\subsubsection*{Acknowledgements.}
The computations were performed on resources provided by the Tensor-GPU project led by Prof. Anne C. Elster through support from The Department of Computer Science and The Faculty of Information Technology and Electrical Engineering, NTNU.
Furthermore, Rudolf Mester acknowledges the support obtained from DNV GL.

\bibliographystyle{splncs04}
\bibliography{bib_files/main,bib_files/image_inpainting,bib_files/partial_convolution,bib_files/gan,bib_files/video_inpainting,bib_files/image_inpainting_trad}

\end{document}